\documentclass[a4paper]{interact}
\usepackage{hyperref}
\usepackage{subcaption}
\usepackage{cite}
\usepackage{soul,color}
\begin{document}
\articletype{Letter}
\title{Multi-head Spatial-Spectral Mamba for Hyperspectral Image Classification}
\author{
\name{Muhammad Ahmad\textsuperscript{a}, Muhammad Hassaan Farooq Butt\textsuperscript{b}, Muhammad Usama\textsuperscript{a}, Hamad Ahmed Altuwaijri\textsuperscript{c}, Manuel Mazzara\textsuperscript{d}, Salvatore Distefano\textsuperscript{d}}
\affil{\textsuperscript{a} M. Ahmad and M. Usama are with the Department of Computer Science, National University of Computer and Emerging Sciences, Islamabad, Chiniot-Faisalabad Campus, Chiniot 35400, Pakistan.; \\ \textsuperscript{b} M. H. F. Butt is with the Institute of Artificial Intelligence, School of Mechanical and Electrical Engineering, Shaoxing University, Shaoxing 312000, China. \\
\textsuperscript{c} H.A. Altuwaijri is with the Department of Geography, College of Humanities and Social Sciences, King Saud University, Riyadh, 11451 Saudi Arabia. \\ 
\textsuperscript{d} M. Mazzara is with the Institute of Software Development and Engineering, Innopolis University, Innopolis, 420500, Russia. \\
\textsuperscript{e} S. Distefano is with  Dipartimento di Matematica e Informatica---MIFT, University of Messina, Messina 98121, Italy.}
}
\maketitle
\begin{abstract}
Spatial-Spectral Mamba (SSM) improves computational efficiency and captures long-range dependencies, addressing Transformer limitations. However, traditional Mamba models overlook rich spectral information in HSIs and struggle with high dimensionality and sequential data. To address these issues, we propose the SSM with multi-head self-attention and token enhancement (MHSSMamba). This model integrates spectral and spatial information by enhancing spectral tokens and using multi-head attention to capture complex relationships between spectral bands and spatial locations. It also manages long-range dependencies and the sequential nature of HSI data, preserving contextual information across spectral bands. MHSSMamba achieved remarkable classification accuracies of 97.62\% on Pavia University, 96.92\% on the University of Houston, 96.85\% on Salinas, and 99.49\% on Wuhan-longKou datasets. The source code is available at \href{https://github.com/MHassaanButt/MHA\_SS\_Mamba}{GitHub}.
\end{abstract}
\section{Introduction}

Hyperspectral Image Classification (HSIC) is essential in various fields due to its capacity to capture detailed spectral information across numerous narrow bands, enabling precise material identification and analysis. This capability has facilitated significant advancements and applications in domains such as remote sensing \cite{ahmad2021hyperspectral}, Earth observation \cite{hong2024multimodal}, urban planning \cite{li2024HD}, agriculture \cite{lu2020recent}, mineral exploration \cite{bedini2017use}, environmental monitoring \cite{stuart2019hyperspectral}, and climate change \cite{pande2023application}. Additionally, HSIC is valuable in less geoscience-related fields such as food processing \cite{khan2021hyperspectral, khan2020hyperspectral}, bakery products \cite{saleem2020prediction}, bloodstain identification \cite{butt2022fast, zulfiqar2021hyperspectral}, and meat processing \cite{ayaz2020hyperspectral, ayaz2020myoglobin}. 

The extensive spectral data in Hyperspectral Images (HSIs) presents both challenges and opportunities for effective classification \cite{hong2024spectralgpt}. The successful application of Neural Networks has bolstered recent progress in HSIC \cite{ahmad2020fast, 10423094, hong2023decoupled, 10409250, 10433668}, with a growing interest in Transformer models to enhance HSI analysis. The Transformer architecture has significantly advanced HSIC compared to traditional deep learning (TDL) methods \cite{yao2023extended, 10399798, 9868046, WANG2024102367, 10123084, 10423821, 10400402, SHU2024107351, ma2024spatial}. Transformers, with their self-attention mechanisms, can capture long-range dependencies within spatial-spectral features, enabling a comprehensive understanding of the intricate relationships between spectral bands and spatial information. The quadratic computational complexity of Transformers presents significant challenges, particularly when dealing with high-dimensional HSI data. This complexity can limit their practical applications. Additionally, Transformers typically require a large number of labeled data for training to achieve high performance; otherwise, they are prone to overfitting.

The emergence of Mamba, a state space model-based (SSM) approach, offers a promising solution to the challenges of high-dimensional HSI data. Mamba can capture long-range dependencies like Transformers but with significantly greater computational efficiency, making them well-suited for processing large datasets without compromising performance \cite{rs16132449, app14135683}. Originally designed for sequence data in natural language processing, Mamba models have proven effective by incorporating time-varying parameters into SSMs, achieving linear time complexity while maintaining strong representational capabilities \cite{yao2024spectralmamba}. This innovative approach has been extended to various visual tasks, including HSIC, where TDL often struggles with the data's high dimensionality and complexity \cite{yang2024hsimamba}.

Mamba's unique cross-scan strategy bridges 1D and 2D sequence scanning, effectively modeling spatial dependencies in multidimensional data. The architecture by Li and Wang et al. \cite{liu2024vision, wang2024s2} includes selective scanning mechanisms and optimizations that enhance training and inference efficiency. Mamba's integration with hybrid models, like combining CNNs with SSMs, offers a robust framework for high-resolution HSIs \cite{zhou2024mamba, he20243dssmamba}. However, Mamba models often neglect the rich spectral information in HSIs, leading to suboptimal performance in tasks requiring spectral feature distinction. Existing models \cite{yao2024spectralmamba} struggle to balance spectral and spatial features, missing critical information. Additionally, Mamba faces challenges in capturing long-range dependencies, significant for HSIs with features spread across different regions, and struggles with the sequential nature of HSI data, losing contextual information across spectral bands \cite{rs16132449}.

In response to the challenges posed by traditional Mamba architecture and the complexities of HSI data, we introduce a multi-head Spatial-Spectral Mamba (MHSSMamba) for HSIC. By building on the Mamba framework, the proposed architecture incorporates cutting-edge token enhancement and multi-head self-attention mechanisms. The MHSSMamba dramatically enhances spectral token representation, leading to superior feature extraction and classification performance. Our key contributions are:

\begin{enumerate}
    \item The Spectral-Spatial token extracts separate spectral and spatial tokens from the input HSI patches, allowing the MHSSMamba model to independently leverage both types of information. This approach potentially enhances feature representation and improves classification performance.

    \item The customized Multi-head self-attention (MHSA) mechanism is specifically designed to process spatial-spectral tokens by projecting queries, keys, and values through dense layers and appropriately reshaping them. This design enables the MHSSMamba to perform attention operations more efficiently and effectively, capturing complex relationships between different spectral bands and spatial locations.

    \item The spectral-spatial feature enhancement module introduces a dual gating mechanism that separately enhances spectral and spatial tokens processed by MHSA using learned gating signals. This dual-gate approach enables the model to adaptively refine features based on both spatial and spectral contexts, potentially increasing the discriminative power of the learned representations.

    \item The State Space Model (SSM) introduces a novel way of capturing temporal dynamics by maintaining and updating state representations through learned transitions and updates. This model component integrates sequential dependencies into the HSI processing pipeline, enhancing the ability to model temporal patterns in time-series HSI data.

\end{enumerate}

In summary, the MHSSMamba presents a comprehensive end-to-end pipeline that integrates token generation, multi-head attention, feature enhancement, and state space modeling. This hybrid architecture combines several advanced techniques into a unified framework, offering an intricate approach to HSIC that leverages both spatial and spectral information as well as temporal dynamics. These contributions underscore the model's innovative techniques for addressing HSIC challenges, providing potential improvements in feature representation, attention mechanisms, and temporal modeling.

\section{Spatial-Spectral Mamba with Multi-Head Self-Attention}
\label{Met}

\begin{figure*}
	\centering
	\includegraphics[width=0.98\textwidth]{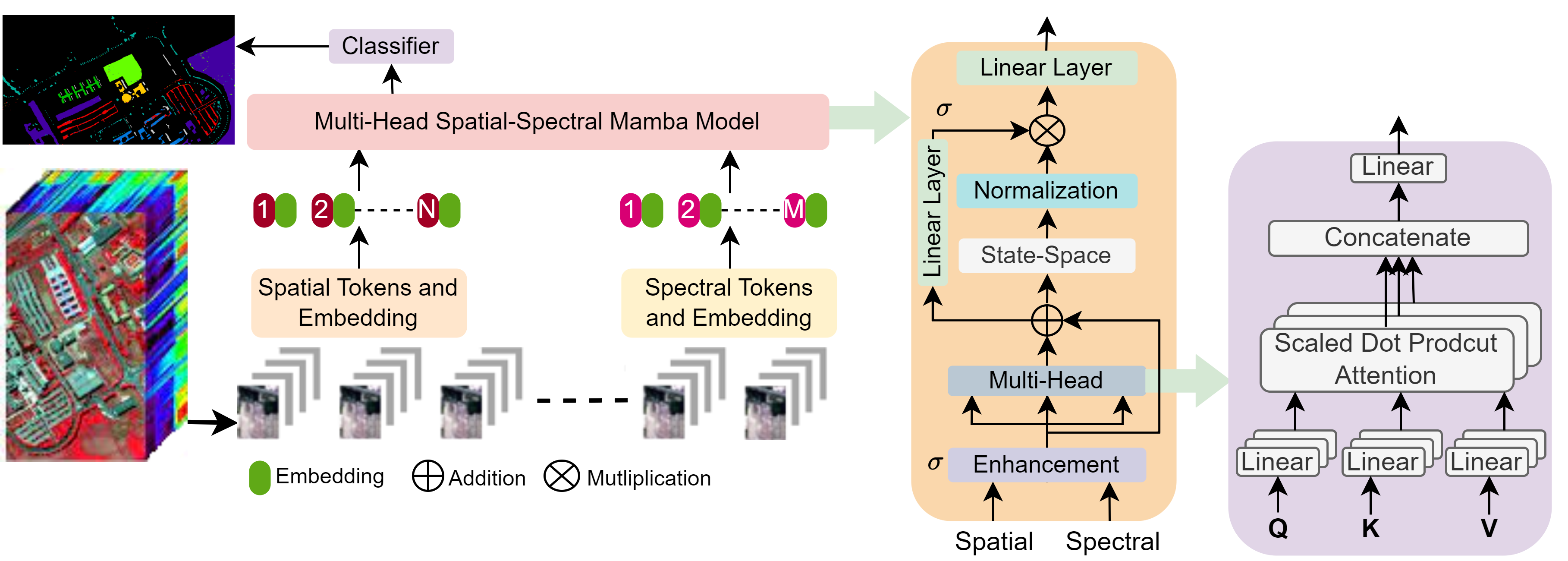}
	\caption{A joint spatial-spectral feature token is first computed from the HSI. These tokens are then encoded in the MHSSMamba model, which includes token enhancement and a multi-head attention module, allowing for a more selective and effective representation of information compared to standard fixed-dimension encodings. The output is subsequently processed through a state-space model, followed by normalization and a linear layer, before being passed to the classification head for ground truth generation.}
	\label{Fig1}
\end{figure*}

Assume the HSI data has a shape of $(H, W, C)$, where $H$ and $W$ are the height and width, and $C$ is the number of bands. Figure \ref{Fig1} provides an overview of the MHSSMamba method. The HSI cube $X$ is divided into overlapping 3D patches as $N = \big(\frac{H}{P} \times \frac{W}{P}\big)$, where $Patch(X) \in \mathcal{R}^{N \times (P \times P \times C)}$, $P$ be the size of patch. These patches are divided into spectral and spatial patches, each undergoing separate processing to generate spectral and spatial tokens.

HSIs have rich spatial and spectral information. By flattening 2D spatial data into spatial tokens $\mathbf{S}$ and 1D spectral data into spectral tokens $\mathbf{F}$, the model captures intricate spatial-spectral relationships. This token representation enables the multi-head attention module to jointly attend to spatial and spectral cues, which is crucial for accurate HSI classification. We generate spatial and spectral tokens $\mathbf{S}$ and $\mathbf{F}$ as $\mathbf{S} = [\mathbf{s}_1, \mathbf{s}_2, ..., \mathbf{s}_C] \in \mathbb{R}^{B \times (HW) \times C}$ and $\mathbf{F} = [\mathbf{f}_1, \mathbf{f}_2, ..., \mathbf{f}_{HW}] \in \mathbb{R}^{B \times (HW) \times C}$. The token enhancement applies a gating mechanism to spatial and spectral tokens, using the center region of the HSI sample as context \cite{rs16132449}. This allows MHSSMamba to adjust the importance of different tokens dynamically, enhancing feature extraction. Given the center tokens, the token enhancements are made as follows:
\begin{equation}
    \widetilde{\mathbf{S}}^{(l)} = \mathbf{S}^{(l)} \odot \sigma(\mathbf{W}_s \mathbf{c} + \mathbf{b}_s)
\end{equation}

\begin{equation}
    \widetilde{\mathbf{F}}^{(l)} = \mathbf{F}^{(l)} \odot \sigma(\mathbf{W}_f \mathbf{c} + \mathbf{b}_f)
\end{equation}
where $\mathbf{S}^{(l)}$ and $\mathbf{F}^{(l)}$ represents the spatial and spectral tokens at layer $l$ of the model. Spatial and spectral tokens capture spatial and spectral features from the HSI samples, respectively. $\mathbf{c}$ denotes the center region of the HSI sample used as context. It helps to provide additional information about the central part of the patch to guide the gating mechanism. $\mathbf{W}_s$ and $\mathbf{W}_f$ are the learned weight parameters used to transform the center context $\mathbf{c}$ for spatial and spectral tokens, respectively. It is a matrix that projects the context into the same space as the spatial and spectral tokens. $\mathbf{b}_s$ and $\mathbf{b}_f$ represent the bias terms added to the transformed context for spatial and spectral tokens. $\sigma$ denotes the sigmoid function, which is used to apply a gating mechanism. The sigmoid function outputs values between 0 and 1, allowing the model to adjust the importance of different tokens dynamically. $\odot$ denotes element-wise multiplication. It is used to apply the gating values (obtained from the sigmoid function) to the spatial and spectral tokens.

The multi-head attention mechanism enables the model to learn diverse and informative feature representations by attending to different parts of the input tokens (both spatial and spectral). By projecting the input tokens into multiple attention heads, the model captures complex dependencies within HSI data, essential for effective feature extraction. The scaled dot-product attention, followed by a softmax operation, dynamically assigns importance weights to the input tokens, enhancing the model's focus on the most relevant features. Let the spatial and spectral tokens $\widetilde{\mathbf{S}}^{(l)}$ and $\widetilde{\mathbf{F}}^{(l)}$ be the inputs, and $Q, K, V$ be the query, key, and value respectively. For each head $i$: $Q_i = \widetilde{\mathbf{S}}^{(l)} W^Q_i$, $K_i = \widetilde{\mathbf{F}}^{(l)} W^K_i$, and $V_i = \widetilde{\mathbf{F}}^{(l)} W^V_i$, where $W^Q_i, W^K_i, W^V_i$ are learned weight matrices. The attention scores are computed as:

\begin{equation}
    A_i = \text{softmax}\left(\frac{\mathbf{Q_i}\mathbf{K_i}^\top}{\sqrt{d_k}}\right)
\end{equation}
where the attention output is $O_i = A_i V_i$ while the concatenated heads are $O = Concat(O_1, O_2, \dots, O_h)$. Given a sequence of enhanced tokens $O = (E_1, E_2, E_3, \dots, E_T)$, the state transition is computed as:

\begin{equation}
    h_t = ReLU(W_{transition}h_{t-1} + W_{update}E_t)
\end{equation}
where $O = (E_1, E_2, E_3, \dots, E_T)$ denotes a sequence of enhanced tokens, where $E_t$ represents the token at the time step $t$, and $T$ is the total number of tokens in the sequence. $h_t$ represents the hidden state at the time step $t$. It captures the context and dependencies from previous tokens and is updated based on the enhanced token $E_t$. $W_{transition}$ is a learned weight matrix used in the transition function. It transforms the previous hidden state $h_{t-1}$ to combine with the current token $E_t$ in the state update process. $W_{update}$ is a learned weight matrix applied to the current enhanced token $E_t$ during the state update process. $ReLU$ denotes the Rectified Linear Unit activation function, which introduces non-linearity into the model by outputting the maximum of 0 and the input value. It is applied to the weighted sum of the previous hidden state and the current token. The final output is obtained by applying a linear classifier on $h_t$:

\begin{equation}
    y = \sigma(h_t W_{classifier})
\end{equation}
where $W_{classifier}$ is a learned weight matrix used in the linear classifier. It projects the hidden state $h_t$ to obtain the final output. $\sigma$ denotes the sigmoid function, which is used to generate probabilities for classification. It outputs values between 0 and 1. $y$ represents the final output of the model, which is obtained by applying the linear classifier to the hidden state $h_t$ and passing it through the sigmoid function.

Integrating spectral-spatial token generation, token enhancement, multi-head attention, and the state space module, MHSSMamba captures complex dependencies in HSI data. By concatenating final spatial and spectral representations and applying a linear classifier, the model leverages complementary information from both modalities for improved classification performance.

\section{Experimental Results and Discussion}
\label{ERD}

The proposed MHSSMamba model is evaluated using several publicly available Hyperspectral datasets including \textbf{The WHU-Hi-LongKou} \cite{8573977, ZHONG2020112012}, \textbf{The University of Pavia (UP)}, \textbf{The Salinas (SA)}, and \textbf{The University of Houston (UH)} datasets. MHSSMamba's weights were initialized randomly and optimized over 50 epochs using the Adam optimizer with a learning rate of 0.001 and softmax loss. The training was done in mini-batches of 256 samples per epoch. The Mamba block's embedding dimensions were set to 64, with 4 heads for multi-head attention and state space dimensions of 128. 

This setup allowed MHSSMamba to learn patterns by adjusting its parameters to minimize loss. Training, validation, and test samples were randomly selected, with various random split percentages tested. The initial patch size was $4 \times 4$, but different sizes were also evaluated, and an optimized set of samples and patch sizes was chosen for comparative results. All experiments were conducted on Google Colab, utilizing a Python 3 notebook with a GPU, 25 GB of RAM, and 358.27 GB of cold storage. 

\begin{figure*}[!hbt]
    \centering
	\begin{subfigure}{0.49\textwidth}
		\includegraphics[width=0.99\textwidth]{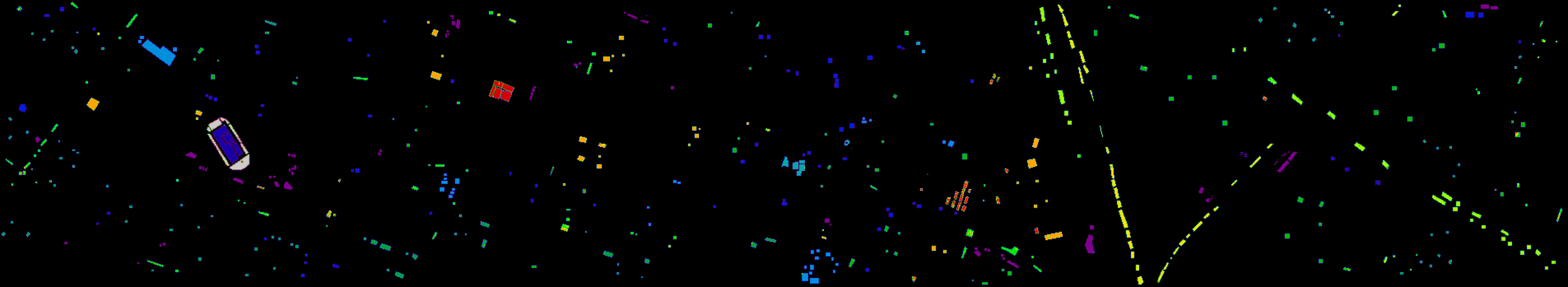}
		\caption{$2 \times 2$} 
		\label{Fig3A}
	\end{subfigure}
	\begin{subfigure}{0.49\textwidth}
		\includegraphics[width=0.99\textwidth]{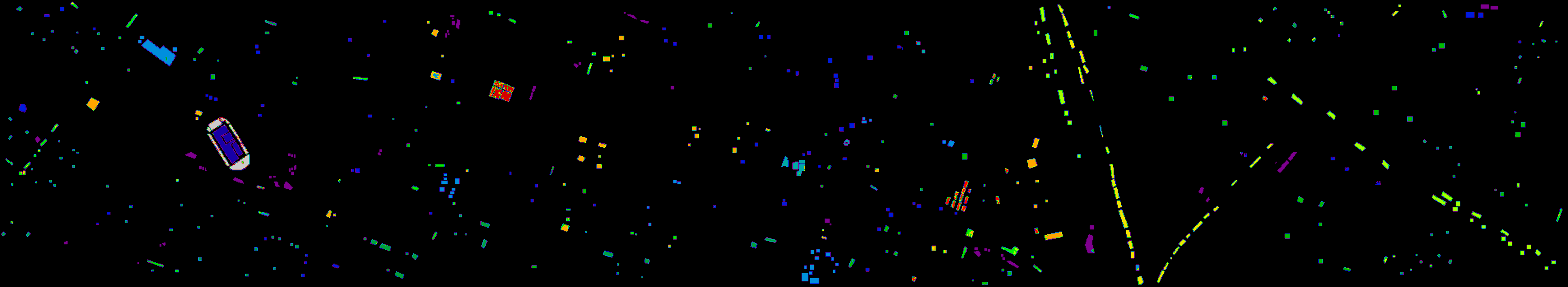}
		\caption{$4 \times 4$}
		\label{Fig3B}
	\end{subfigure}
	\begin{subfigure}{0.49\textwidth}
		\includegraphics[width=0.99\textwidth]{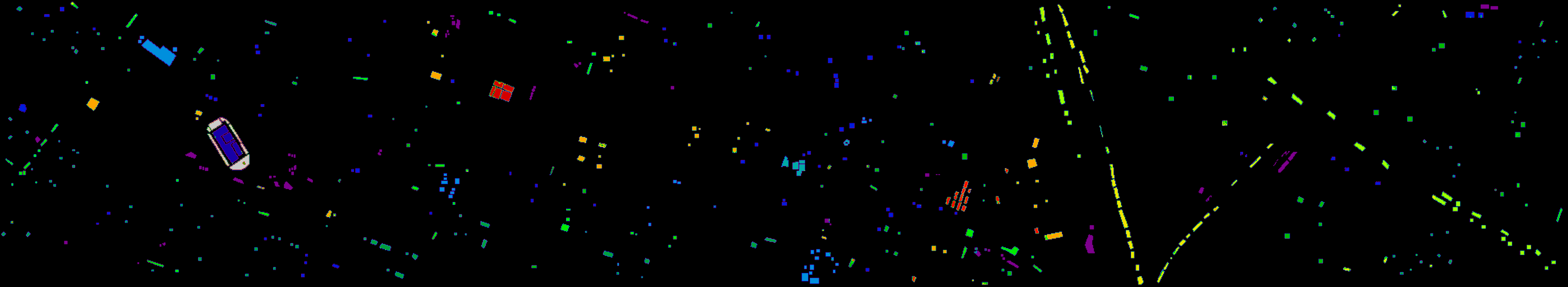}
		\caption{$6 \times 6$}
		\label{Fig3C}
	\end{subfigure}
	\begin{subfigure}{0.49\textwidth}
		\includegraphics[width=0.99\textwidth]{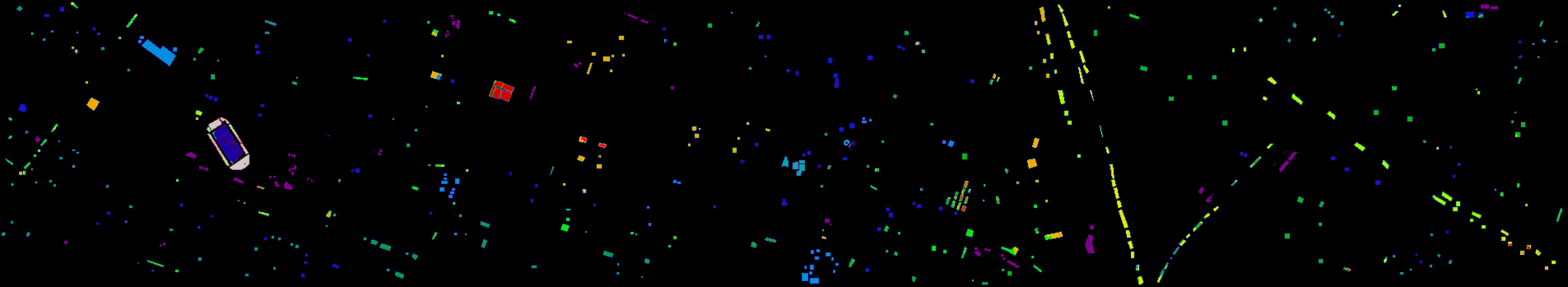}
		\caption{$8 \times 8$}
		\label{Fig3D}
	\end{subfigure}
	\begin{subfigure}{0.49\textwidth}
		\includegraphics[width=0.99\textwidth]{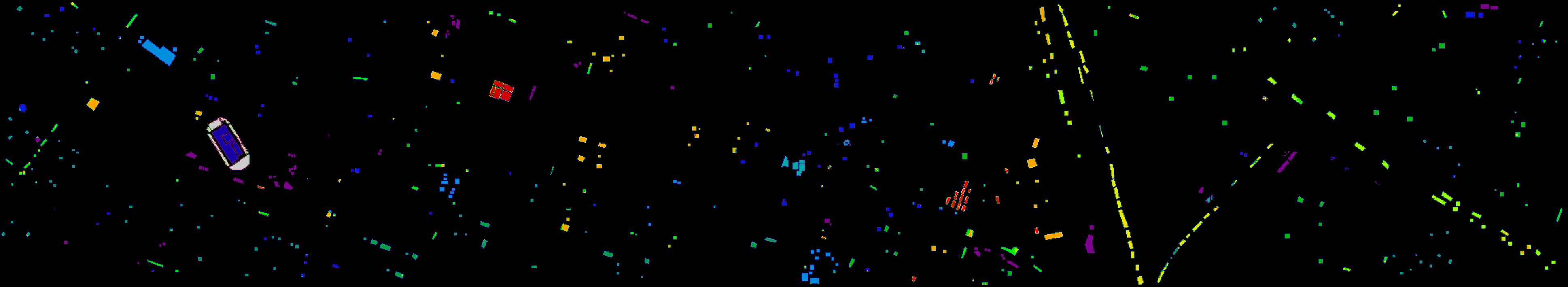}
		\caption{$10 \times 10$}
		\label{Fig3E}
	\end{subfigure} 
\caption{Qualitative results of the University of Houston Dataset.}
\label{Fig3}
\end{figure*}
\begin{figure}[!hbt]
    \centering
	\begin{subfigure}{0.18\textwidth}
		\includegraphics[width=0.99\textwidth]{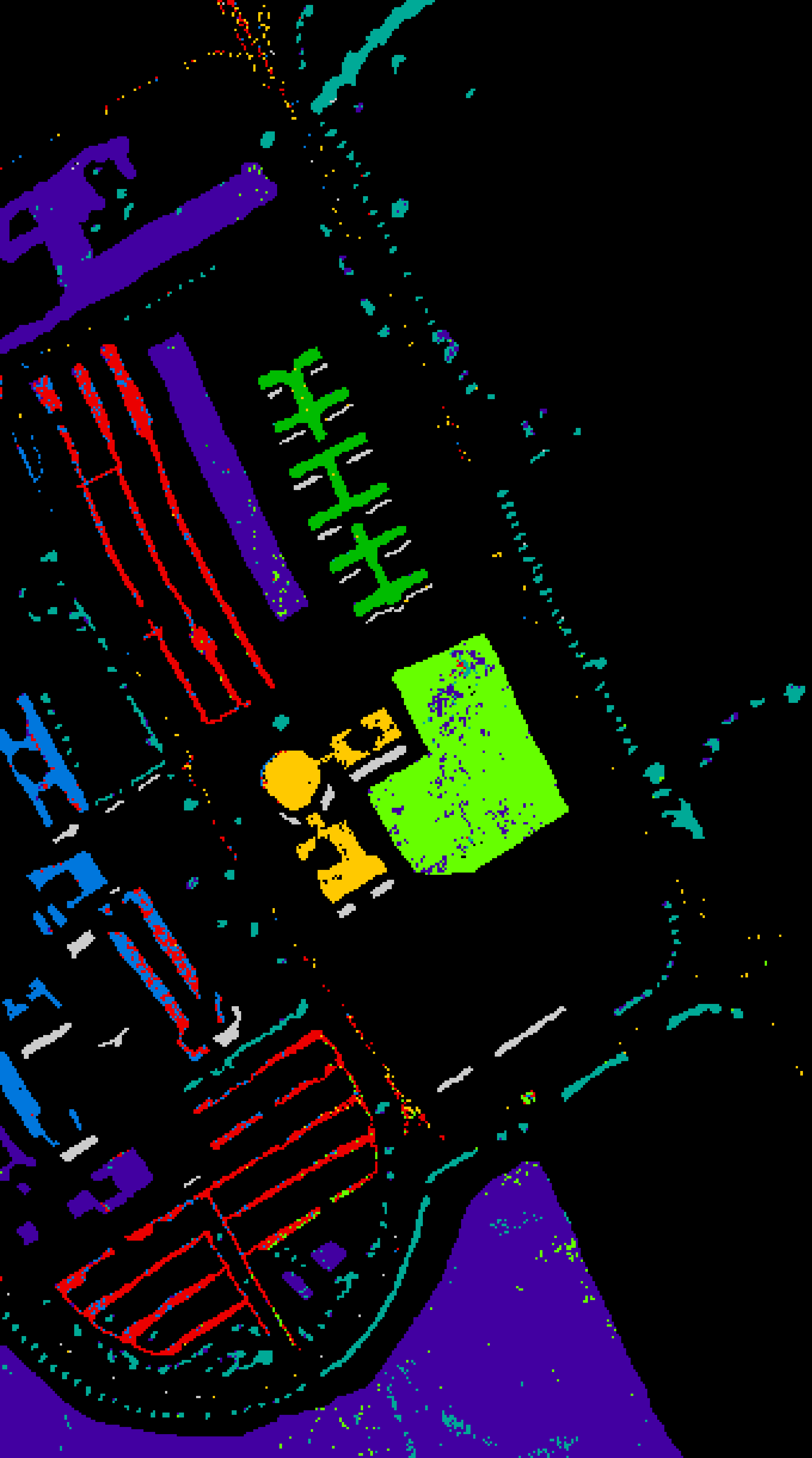}
		\caption{$2 \times 2$} 
		\label{Fig4A}
	\end{subfigure}
	\begin{subfigure}{0.18\textwidth}
		\includegraphics[width=0.99\textwidth]{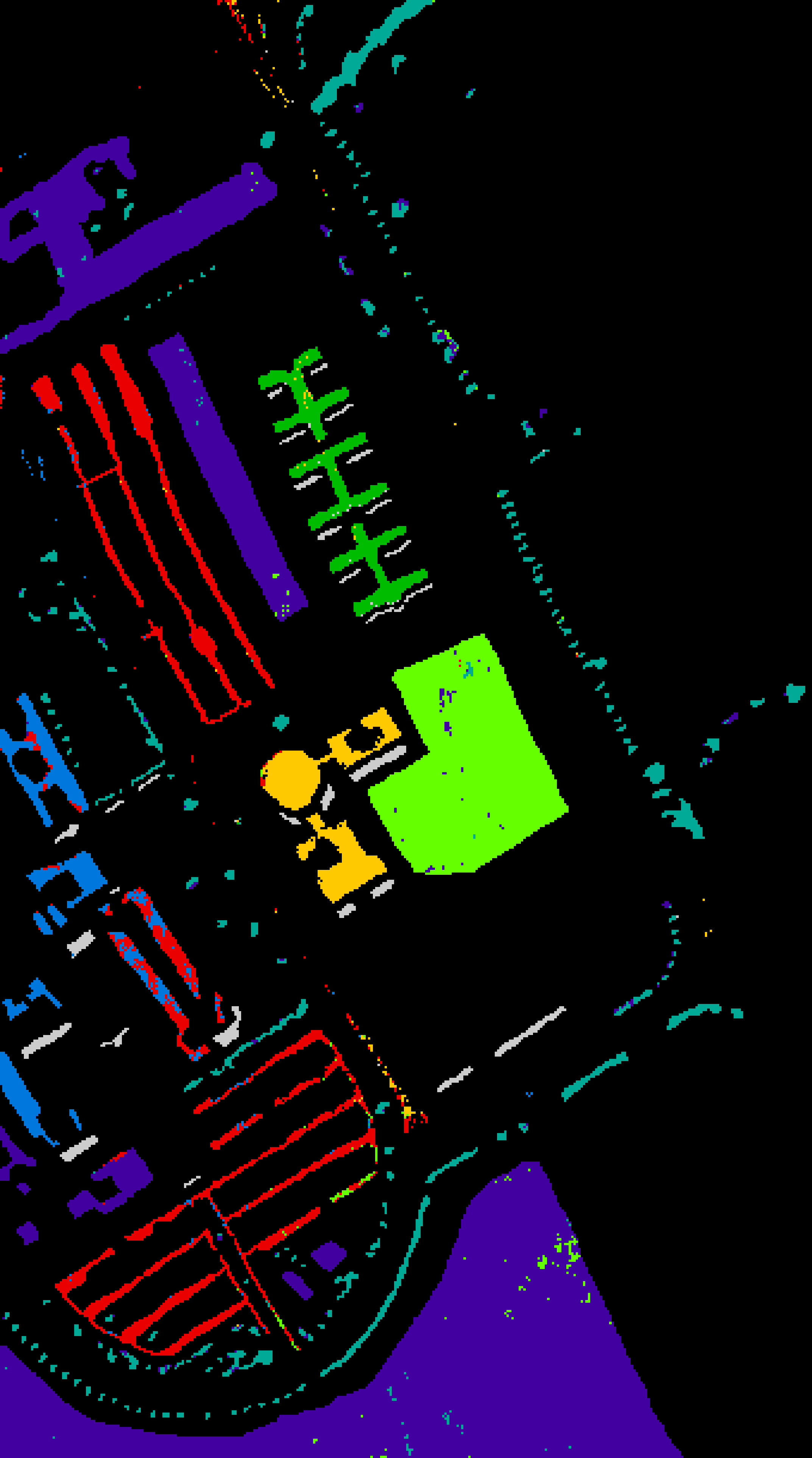}
		\caption{$4 \times 4$}
		\label{Fig4B}
	\end{subfigure}
	\begin{subfigure}{0.18\textwidth}
		\includegraphics[width=0.99\textwidth]{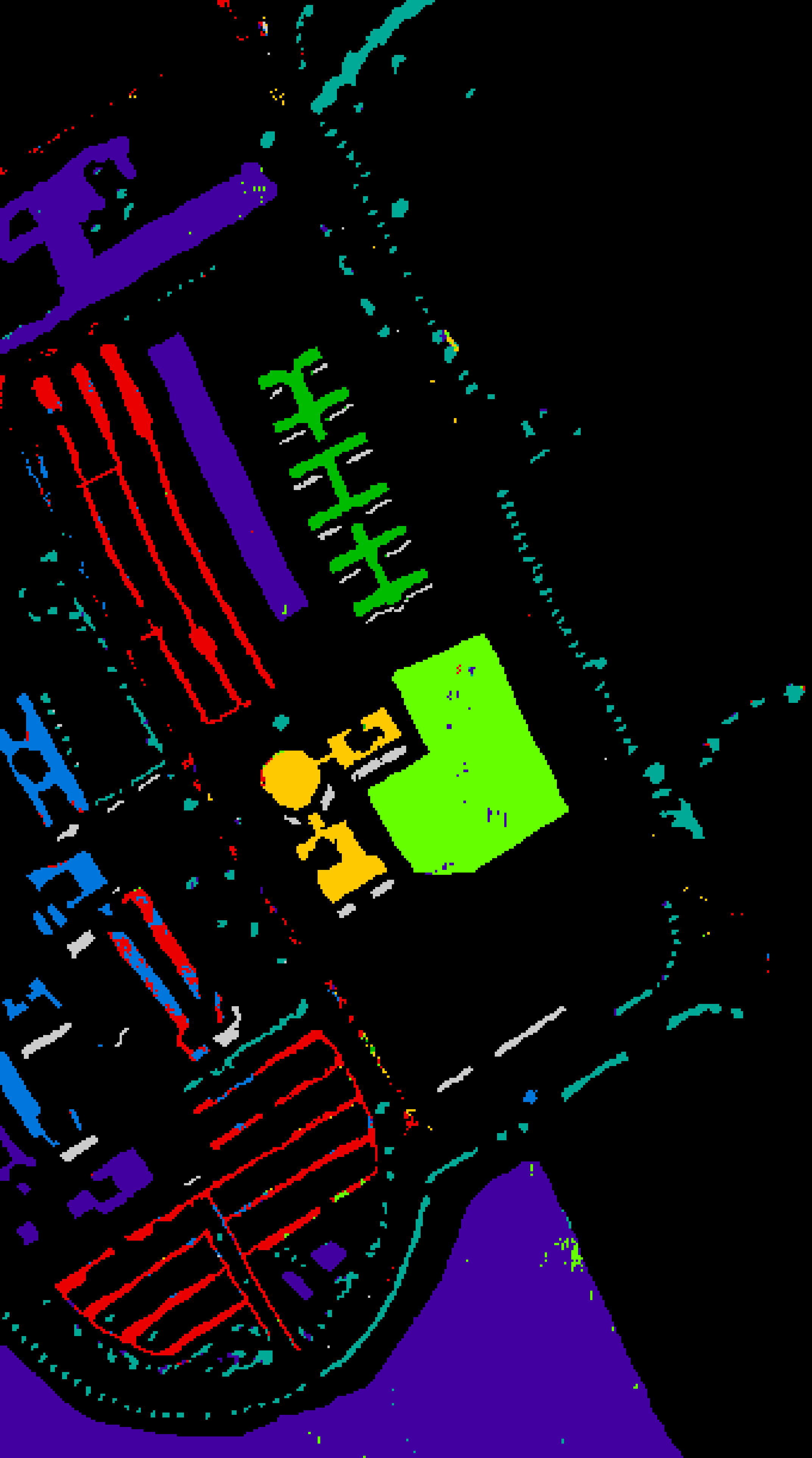}
		\caption{$6 \times 6$}
		\label{Fig4C}
	\end{subfigure}
	\begin{subfigure}{0.18\textwidth}
		\includegraphics[width=0.99\textwidth]{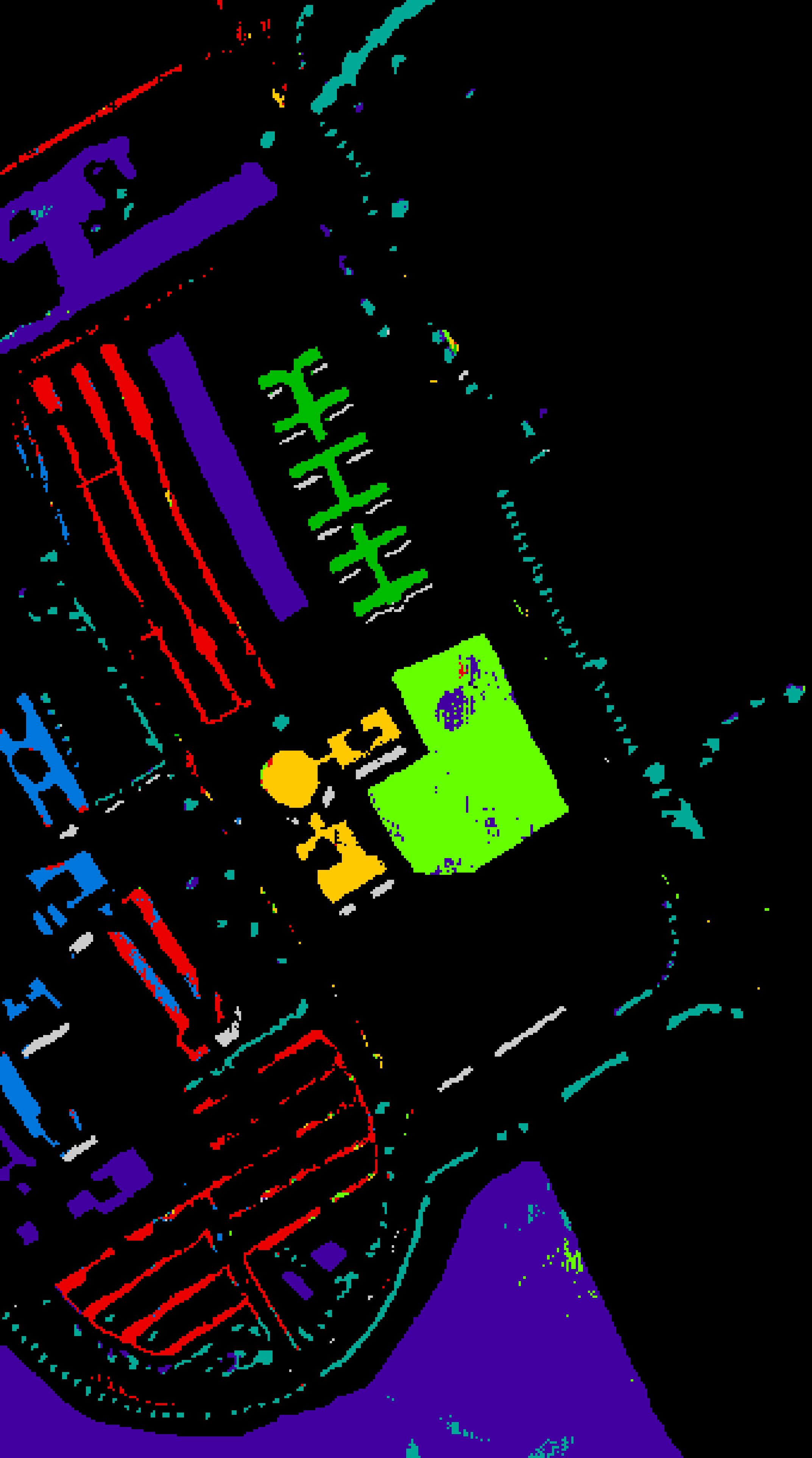}
		\caption{$8 \times 8$}
		\label{Fig4D}
	\end{subfigure}
	\begin{subfigure}{0.18\textwidth}
		\includegraphics[width=0.99\textwidth]{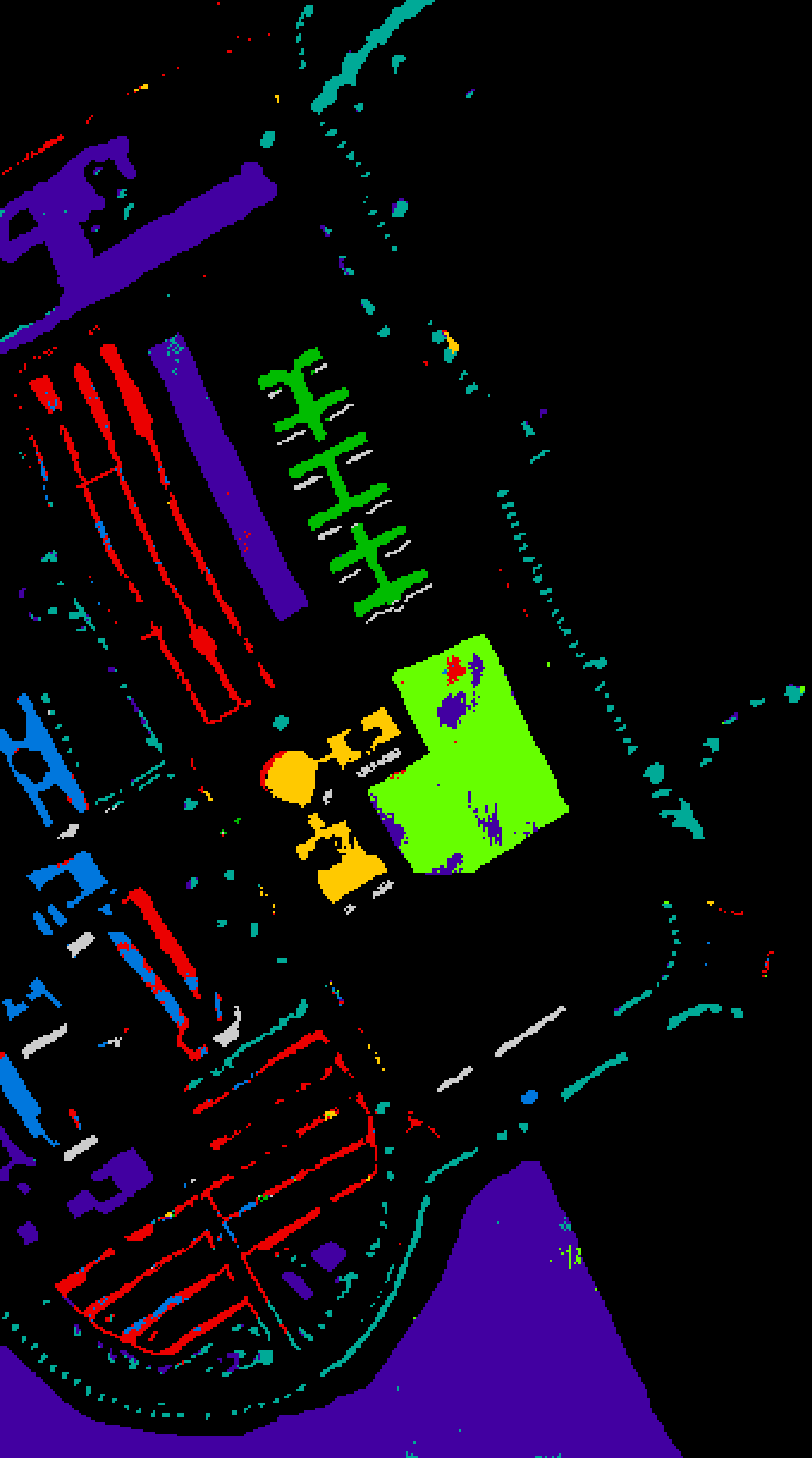}
		\caption{$10 \times 10$}
		\label{Fig4E}
	\end{subfigure} 
\caption{Qualitative results of the Pavia University Dataset.}
\label{Fig4}
\end{figure}
\begin{figure}[!hbt]
    \centering
	\begin{subfigure}{0.18\textwidth}
		\includegraphics[width=0.99\textwidth]{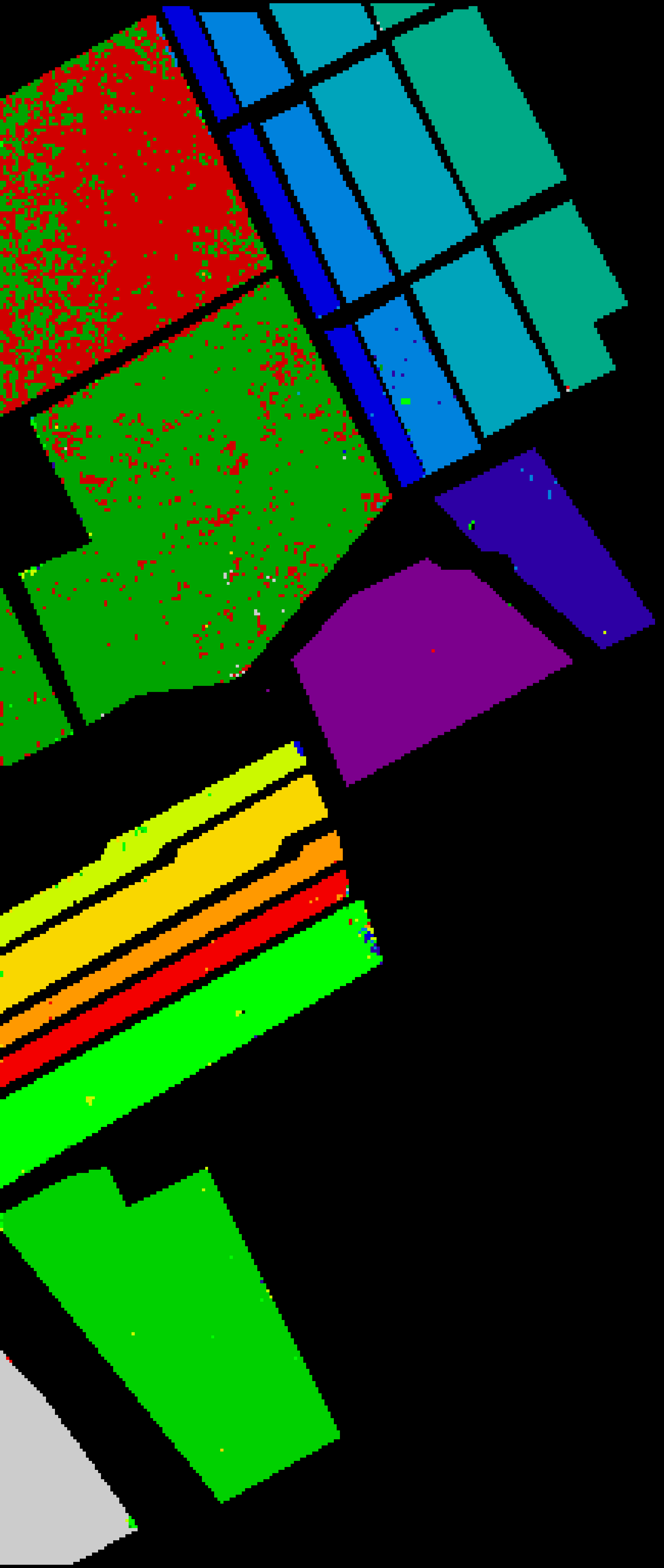}
		\caption{$2 \times 2$} 
		\label{Fig5A}
	\end{subfigure}
	\begin{subfigure}{0.18\textwidth}
		\includegraphics[width=0.99\textwidth]{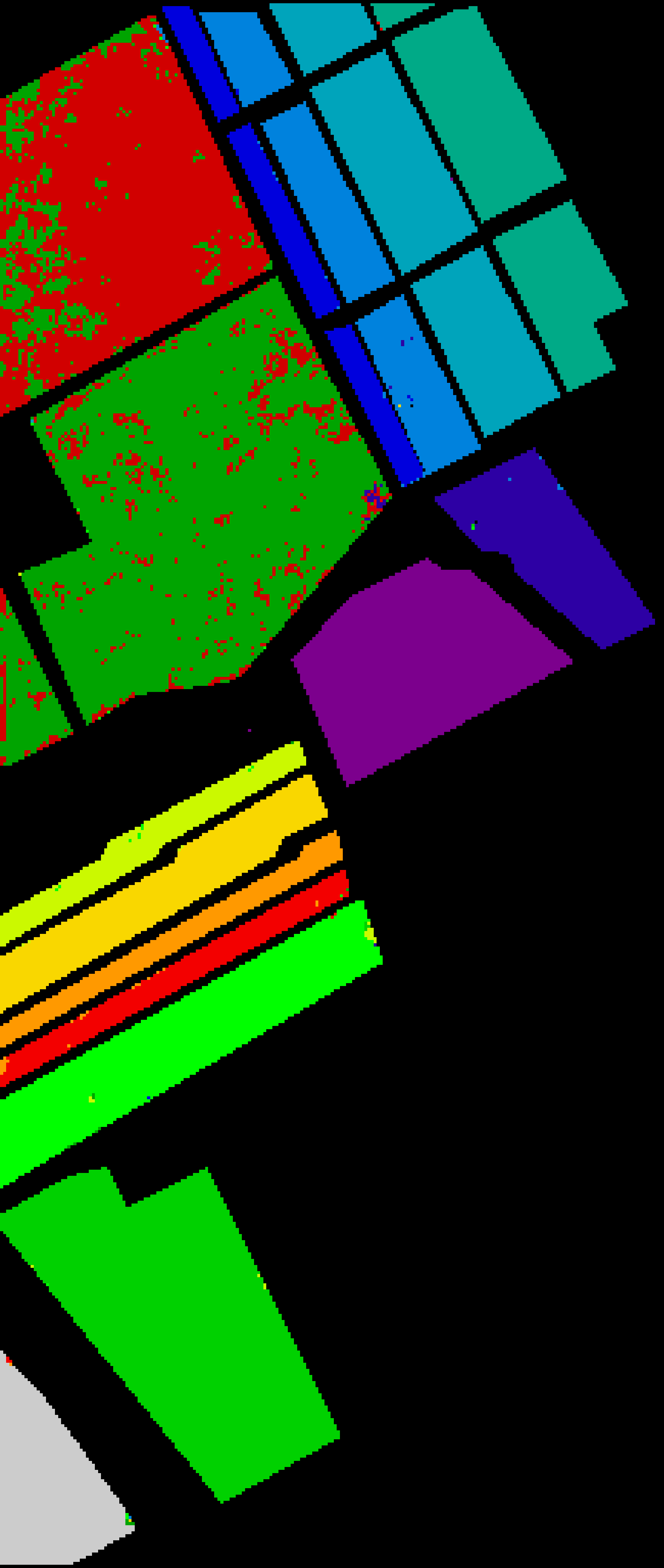}
		\caption{$4 \times 4$}
		\label{Fig5B}
	\end{subfigure}
	\begin{subfigure}{0.18\textwidth}
		\includegraphics[width=0.99\textwidth]{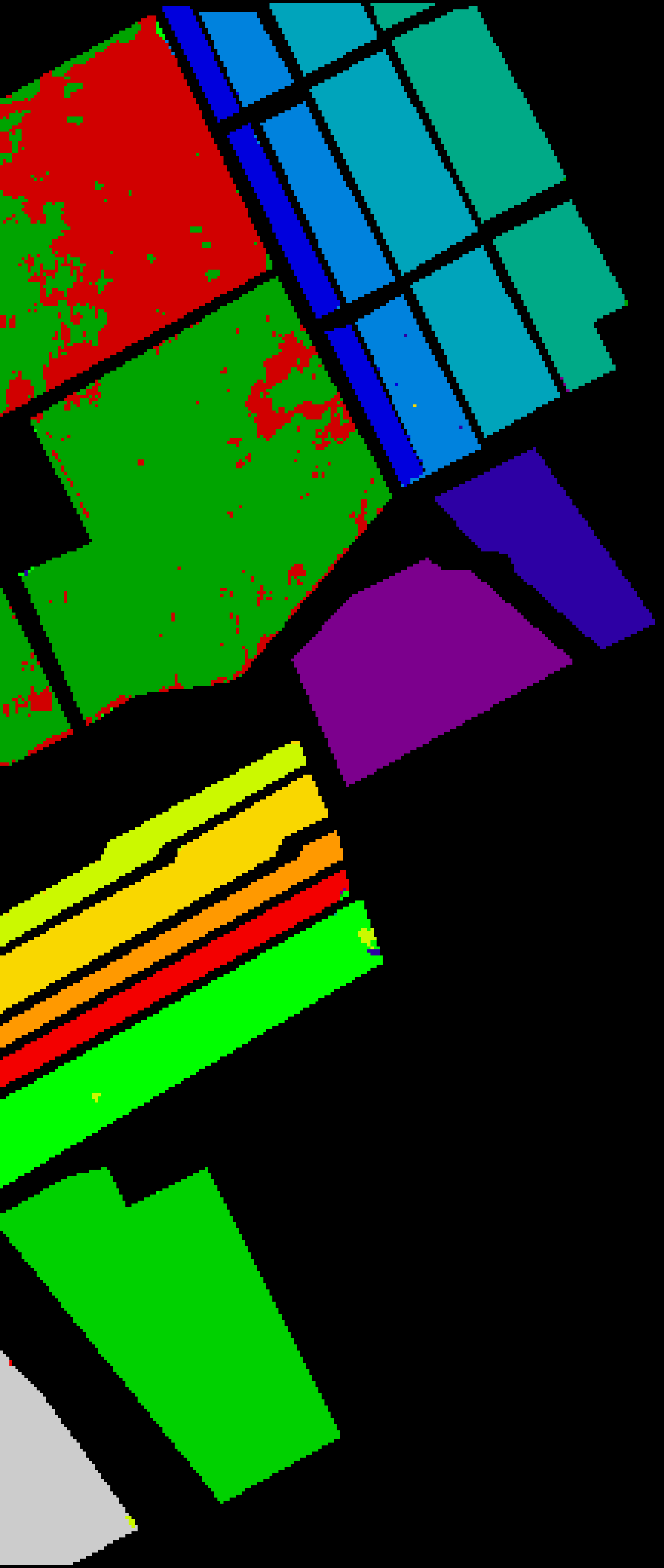}
		\caption{$6 \times 6$}
		\label{Fig5C}
	\end{subfigure}
	\begin{subfigure}{0.18\textwidth}
		\includegraphics[width=0.99\textwidth]{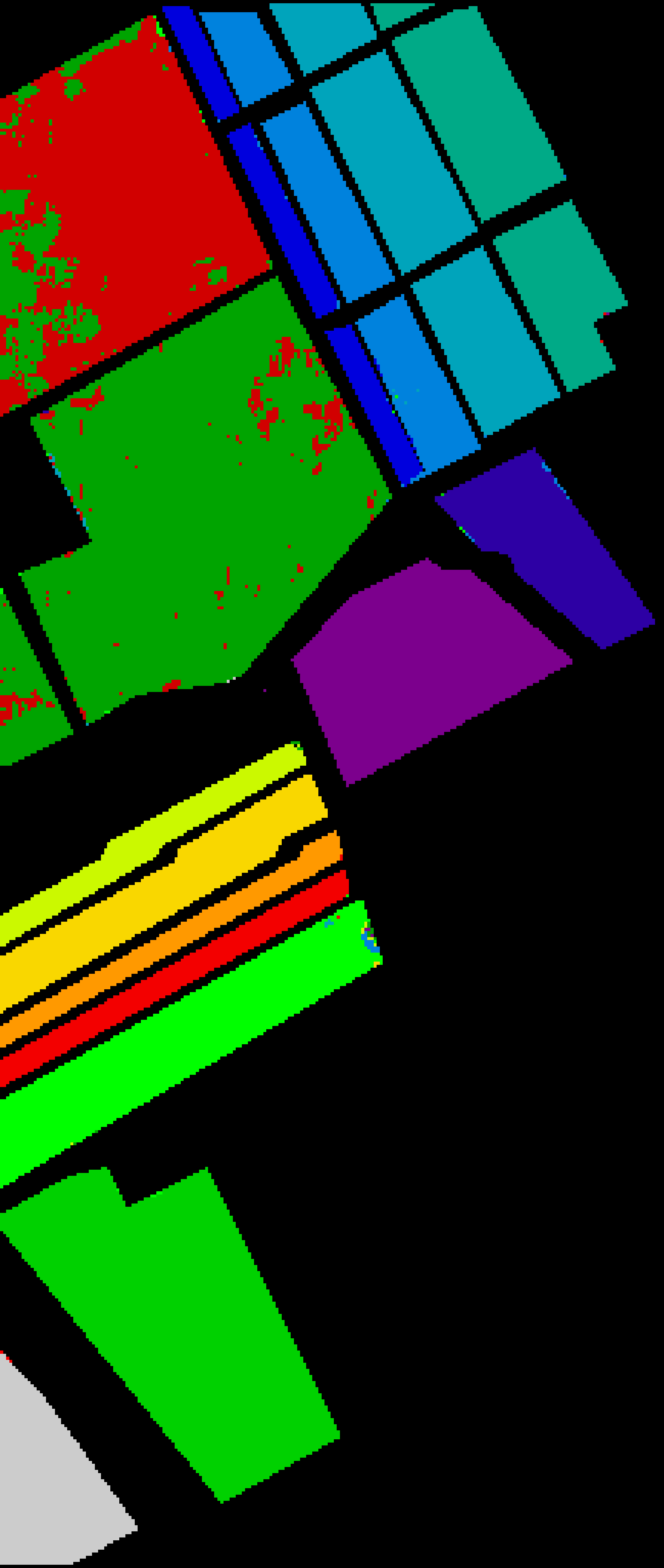}
		\caption{$8 \times 8$}
		\label{Fig5D}
	\end{subfigure}
	\begin{subfigure}{0.18\textwidth}
		\includegraphics[width=0.99\textwidth]{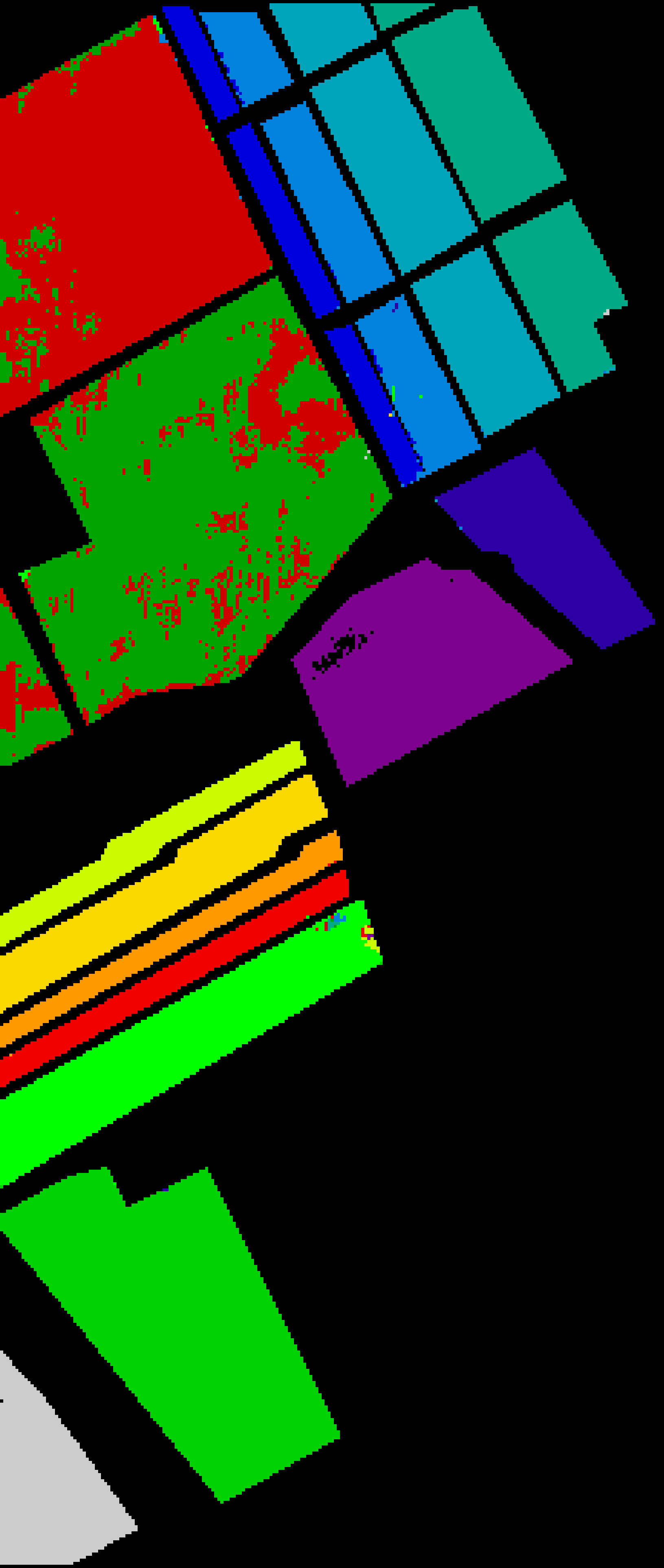}
		\caption{$10 \times 10$}
		\label{Fig5E}
	\end{subfigure} 
\caption{Qualitative results of the Salinas Dataset.}
\label{Fig5}
\end{figure}
\begin{figure}[!hbt]
    \centering
	\begin{subfigure}{0.18\textwidth}
		\includegraphics[width=0.99\textwidth]{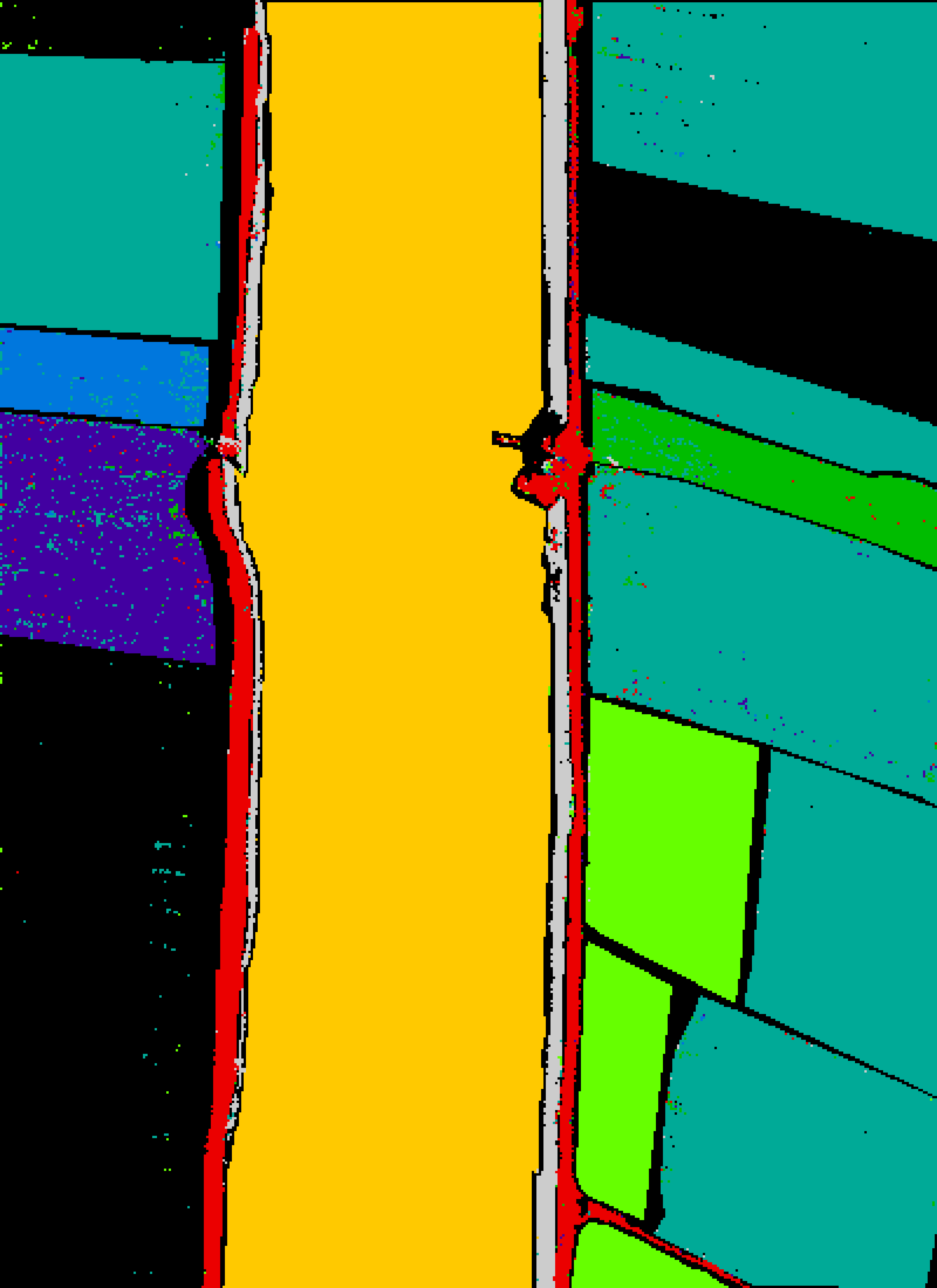}
		\caption{$2 \times 2$} 
		\label{Fig6A}
	\end{subfigure}
	\begin{subfigure}{0.18\textwidth}
		\includegraphics[width=0.99\textwidth]{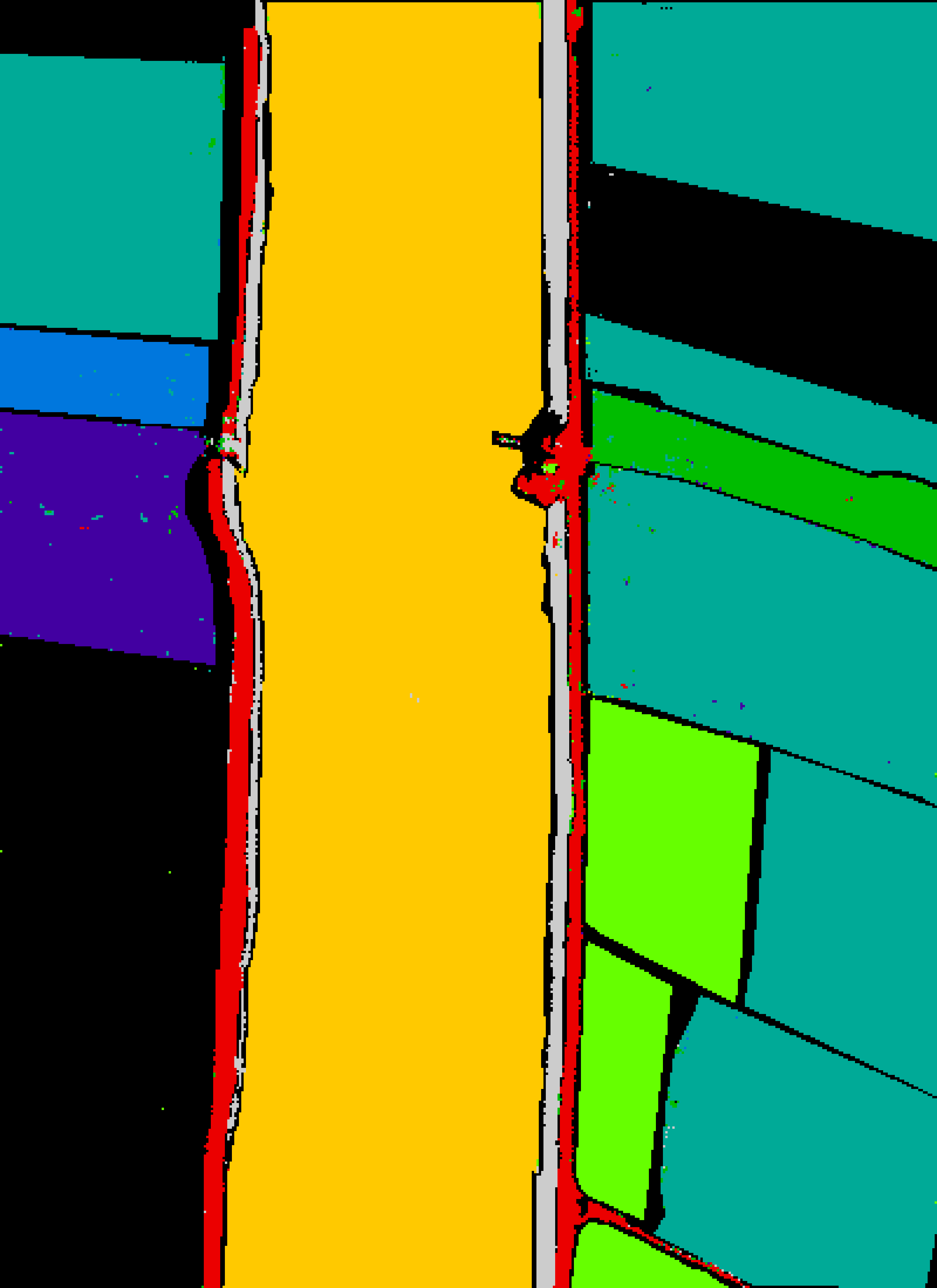}
		\caption{$4 \times 4$}
		\label{Fig6B}
	\end{subfigure}
	\begin{subfigure}{0.18\textwidth}
		\includegraphics[width=0.99\textwidth]{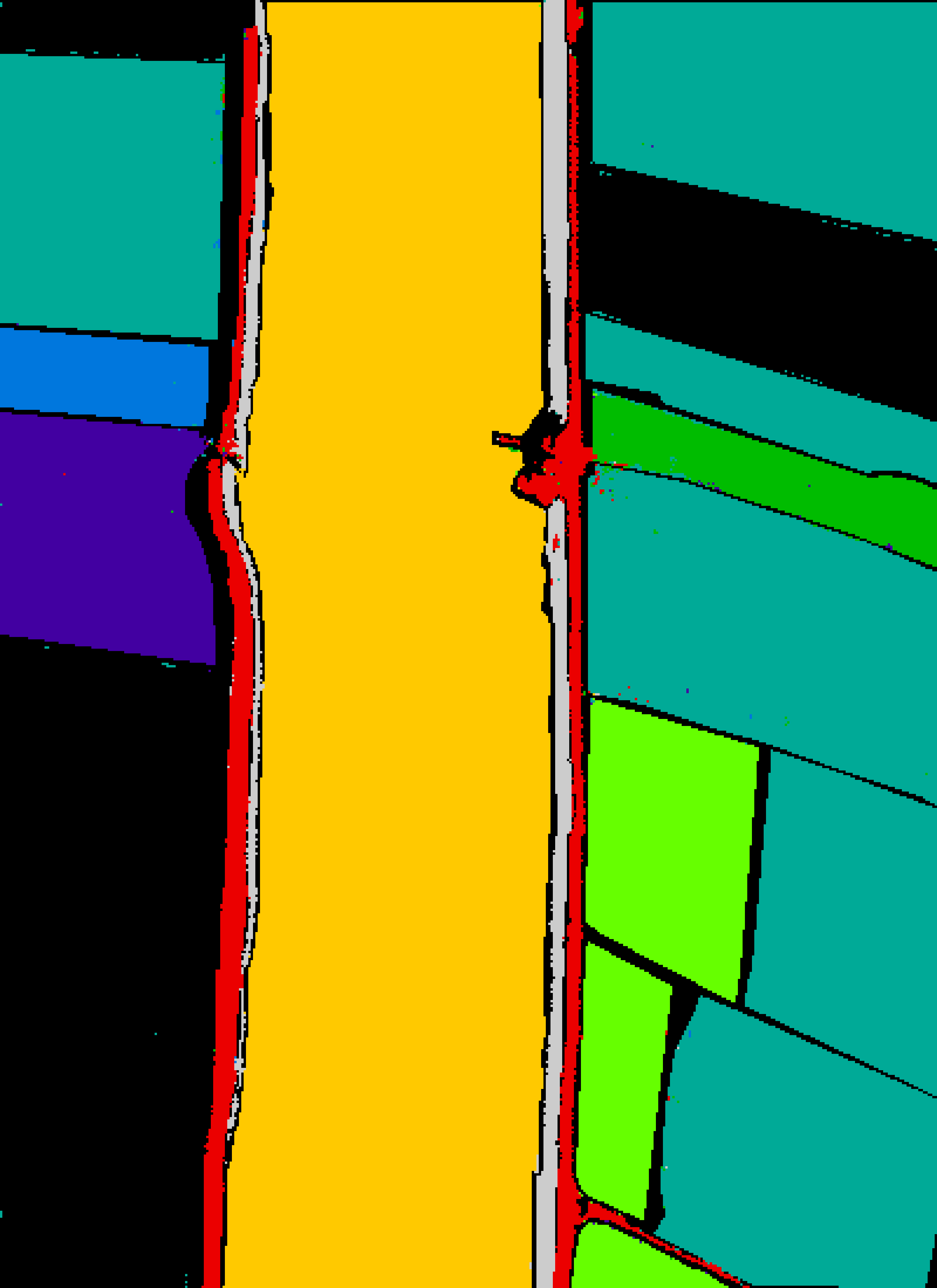}
		\caption{$6 \times 6$}
		\label{Fig6C}
	\end{subfigure}
	\begin{subfigure}{0.18\textwidth}
		\includegraphics[width=0.99\textwidth]{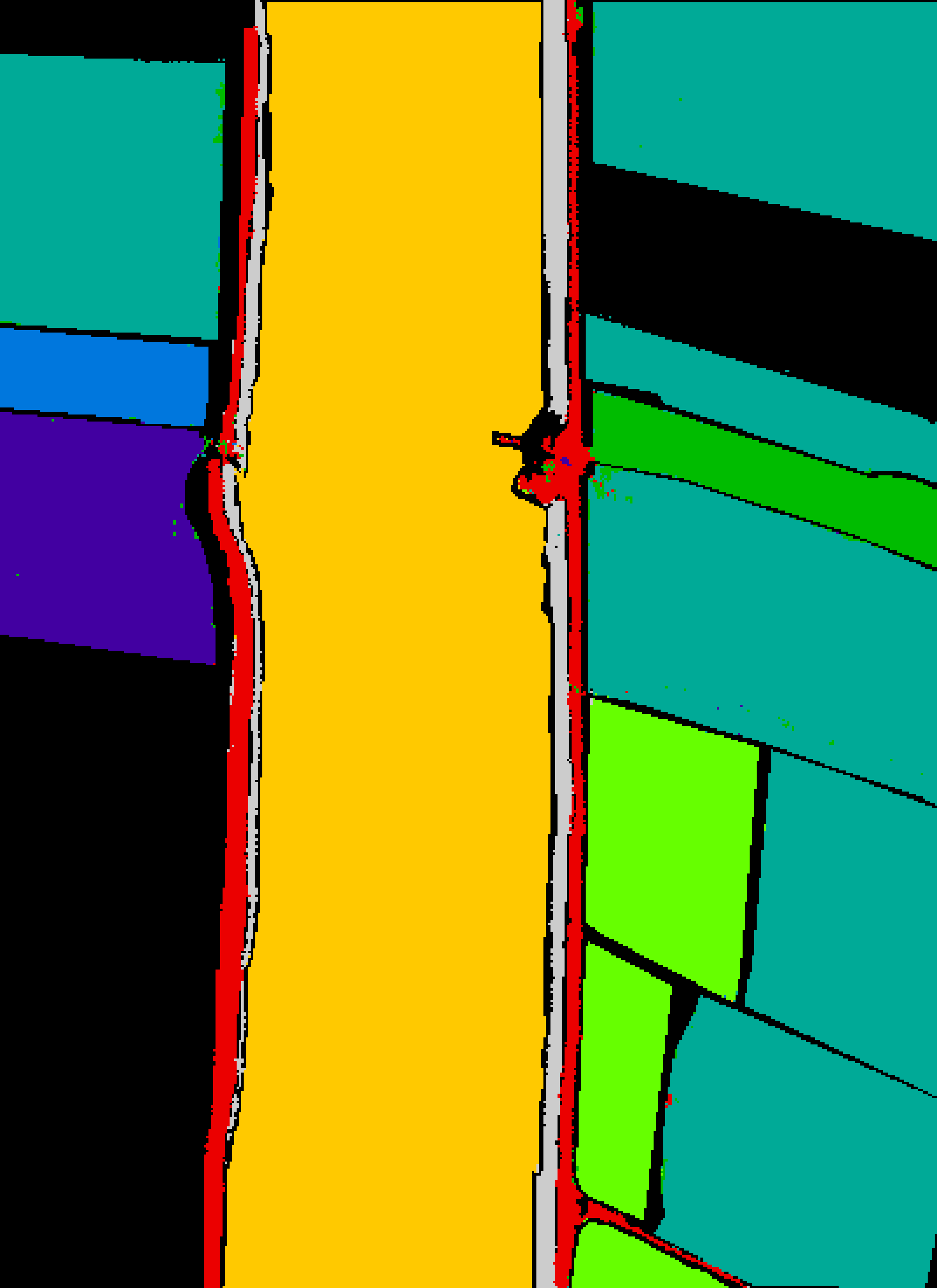}
		\caption{$8 \times 8$}
		\label{Fig6D}
	\end{subfigure}
	\begin{subfigure}{0.18\textwidth}
		\includegraphics[width=0.99\textwidth]{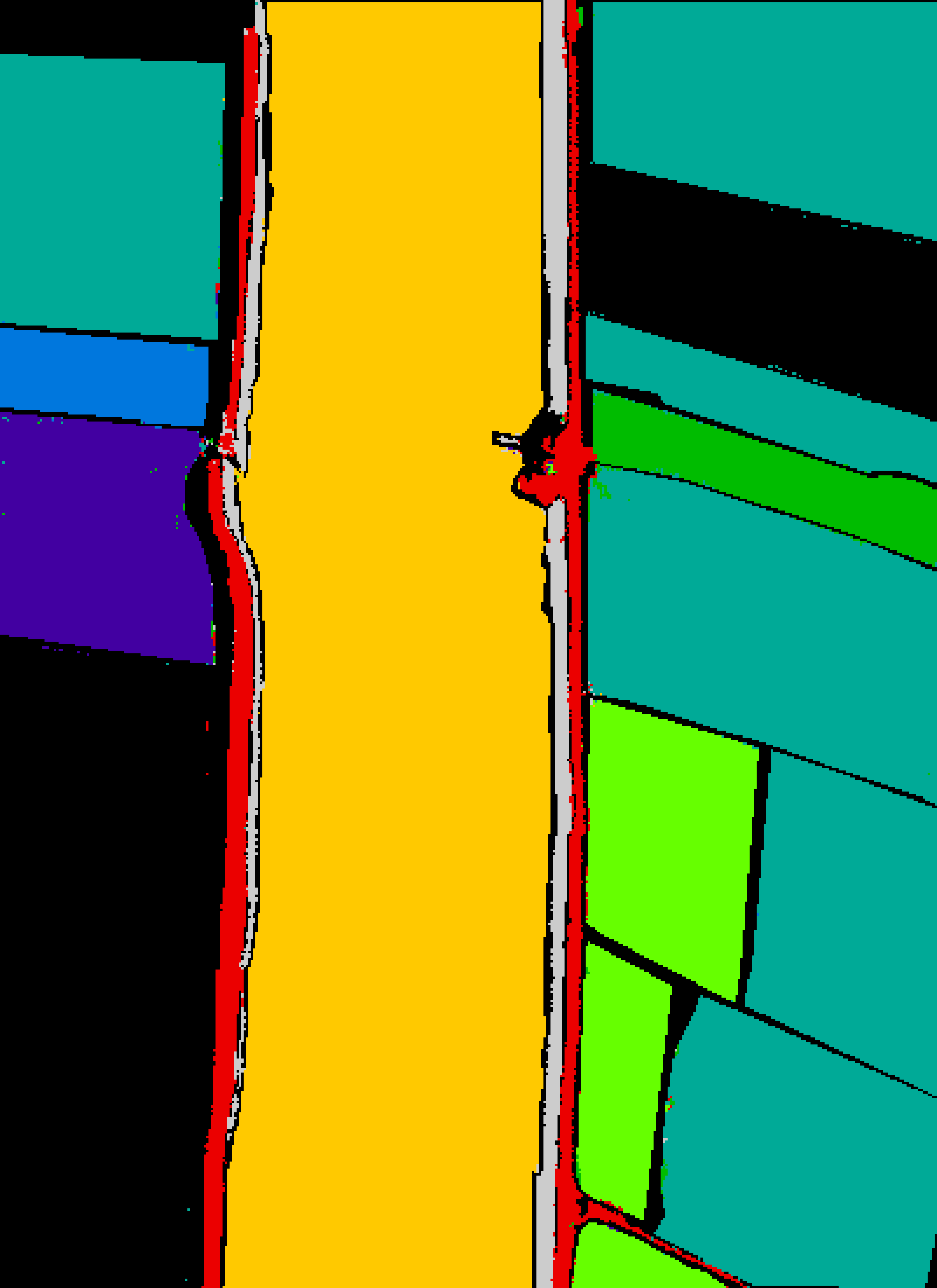}
		\caption{$10 \times 10$}
		\label{Fig6E}
	\end{subfigure} 
\caption{Qualitative results of the WHU-Hi-LongKou Dataset.}
\label{Fig6}
\end{figure}

Table \ref{TabB} presents MHSSMamba's performance with different training and test percentages using a $4 \times 4$ patch size and various patch sizes. Larger patches capture minute details and local patterns but are prone to noise and overfitting, especially with smaller samples. Smaller patches encapsulate global features and contextual information, enhancing robustness against noise. As sample sizes increase, MHSSMamba's robustness and generalization improve. Computational performance is influenced by internet speed and available RAM. Figures \ref{Fig3}, \ref{Fig4}, \ref{Fig5}, and \ref{Fig6} show qualitative results, with quantitative results in Table \ref{TabB}.

\begin{table}[!hbt]
\footnotesize
    \centering
    \caption{The performance of the proposed model is evaluated using various patch sizes and different test splits. For all test splits, a $4 \times 4$ patch size is used over 50 epochs with a batch size of 256. For all patch sizes, 10\% of the data is allocated for training and validation samples.}
    \resizebox{\columnwidth}{!}{\begin{tabular}{c|cccc||c|ccccc} \hline 
        \textbf{Tr/Va/Te(\%)} & \textbf{AA} & \textbf{OA} & \textbf{$\kappa$} & \textbf{Train Time (s)} & \textbf{Patch} & \textbf{AA} & \textbf{OA} & \textbf{$\kappa$} & \textbf{Train Time (s)} \\ \hline
        
        \multicolumn{10}{c}{\textbf{WHU-Hi-LongKou}} \\ \hline 
        5/5/90 & 98.41 & 99.26 & 99.02 & 43.78 & $2 \times 2$ & 96.36 & 98.72 & 98.32 & 50.72 \\ \hline 
        10/10/80 & 98.28 & 99.35 & 99.14 & 78.08 & $4 \times 4$ & 98.74 & 99.53 & 99.39 & 77.46  \\ \hline 
        15/15/70 & 99.31 & 99.70 & 99.60 & 115.60& $6 \times 6$ & 99.12 & 99.68 & 99.58 & 136.40\\ \hline 
        20/20/60 & 99.31 & 99.69 & 99.60 & 168.82 & $8 \times 8$ & 99.23 & 99.63 & 99.52 & 246.62\\ \hline 
        25/25/50 & 98.84 & 99.65 & 99.55 & 192.55 & $10 \times 10$ & 98.89&99.56&99.42&357.10\\ \hline                

        \multicolumn{10}{c}{\textbf{Pavia University}} \\ \hline 
        5/5/90 & 75.15 & 85.70 & 80.59 & 22.33 & $2 \times 2$ & 91.47 & 93.68 & 90.88 & 22.09\\ \hline 
        10/10/80 & 94.33 & 95.53 & 94.04 & 42.74 & $4 \times 4$ & 93.68 & 95.96 & 94.64 & 27.88 \\ \hline 
        15/15/70 & 95.75 & 96.60 & 95.48 & 38.94 & $6 \times 6$ & 94.46 & 96.40 & 95.23 & 85.37\\ \hline 
        20/20/60 & 96.36 & 97.51 & 96.70 & 84.58 & $8 \times 8$ & 89.44 & 93.08 & 90.80 & 70.25\\ \hline 
        25/25/50 & 96.41 & 97.62 & 96.85 & 204.22 & $10 \times 10$ & 88.39 & 93.11 & 90.81 & 143.55\\ \hline  
        
        \multicolumn{10}{c}{\textbf{Salinas}} \\ \hline
        5/5/90 & 97.31 & 94.39 & 93.75 & 18.09 & $2 \times 2$ & 97.12 & 94.18 & 93.51 &  20.96\\ \hline 
        10/10/80 & 97.44 & 94.73 & 94.11 & 42.84 & $4 \times 4$ & 97.79 &95.28&94.75& 33.68\\ \hline 
        15/15/70 & 97.87 & 95.52 & 95.01 & 83.51 & $6 \times 6$ & 97.81 & 95.12& 94.56 & 83.53\\ \hline 
        20/20/60 & 98.21 & 96.27 & 95.84 & 83.53 & $8 \times 8$ & 98.26 &96.48&96.08&143.56\\ \hline 
        25/25/50 & 98.55 & 97.00 & 96.66 & 73.99 & $10 \times 10$ &97.94&94.90&94.33&144.19\\ \hline  
        
        \multicolumn{10}{c}{\textbf{University of Houston}} \\ \hline
        5/5/90 & 90.87 & 91.60 & 90.92 & 11.90 &  $2 \times 2$ & 94.10&94.55&94.10& 8.11\\ \hline 
        10/10/80 & 90.70 & 90.71 & 89.95 & 12.54 & $4 \times 4$ &  92.24 &93.28&92.73& 11.95\\ \hline 
        15/15/70 & 95.42 & 95.40 & 95.03 & 15.70 & $6 \times 6$ & 93.43 &93.42&92.88&19.56\\ \hline 
        20/20/60 & 96.61 & 96.82 & 96.57 & 23.08 & $8 \times 8$ & 94.04 &93.85&93.35&27.48\\ \hline 
        25/25/50 & 96.73 & 96.92 & 96.67 & 42.58 & $10 \times 10$ &92.37&91.09&90.37&42.60\\ \hline  
    \end{tabular}}
    \label{TabB}
\end{table}

\subsection{Comparative Methods}

To demonstrate the effectiveness of MHSSMamba, various HSIC methods were selected for comparison: S3L: Spectrum Transformer for Self-Supervised Learning in HSIC \cite{rs16060970}, RIAN: Rotation-Invariant Attention Network for HSIC \cite{9785505}, CAT: Center Attention Transformer With Stratified Spatial–Spectral Token for HSIC \cite{10463068}, SPRLT: Local Transformer With Spatial Partition Restore for HSIC \cite{9772356}, and CMT: A Center-Masked Transformer for HSIC \cite{10443948}. DBDA \cite{rs12030582}, an advanced CNN model with a double-branch dual-attention mechanism, serves as a benchmark against Transformer-based approaches. MSSG \cite{9547387} employs a super-pixel structured graph U-Net to learn multiscale features. SSFTT \cite{9684381} is a spatial-spectral Transformer with a novel tokenization approach and CNN-generated local features. LSFAT \cite{9864609} aggregates local semantic features to learn multiscale features effectively. CT-Mixer \cite{9903640} combines CNN and Transformer frameworks. SS-Mamba \cite{rs16132449} includes a spectral-spatial token generation module with stacked spectral-spatial Mamba blocks.

\textbf{Evaluation.} To assess the classification performance of the MHSSMamba model, we used three common evaluation metrics: Overall Accuracy (OA), Average Accuracy (AA), and kappa coefficient ($\kappa$). All methods were tested using optimal experimental settings or re-implemented with their official code where applicable.

\begin{table}[!hbt]
\centering
\caption{University of Houston Classification Results over several SOTA methods.}
\resizebox{\columnwidth}{!}{\begin{tabular}{c|cccccccccccc} \hline 
    \textbf{Class} & RIAN & CAT & S3L & SPRLT & CMT & MSSG & DBDA & SSFTT & LSFAT & CT-Mixer & SS-Mamba & \textbf{Proposed} \\ \hline 
    Grass-Heathy & 83.00 & 88.13 & 81.23 & 80.51 & 79.47 & 93.44 & 93.48 & 95.36 & 92.67 & 90.90 & 92.88 & \textbf{97.60} \\ \hline
    Grass-Stressed & 83.74 & 90.04 & 87.59 & 67.45 & 79.08 & 95.37 & 95.10 & 98.58 & 97.57 & 95.11 & \textbf{95.99} & 99.84 \\ \hline
    Grass-Synthetic & 89.70 & \textbf{100.0} & 88.19 & 99.12 & \textbf{100.0} & 98.67 & \textbf{100.0} & 99.20 & 98.17 & 97.10 & \textbf{100.0} & 99.13 \\ \hline
    Tree & 92.57 & 89.49 & 77.05 & 97.39 & 98.51 & 97.69 & 97.13 & 97.82 & 96.58 & 92.27 & \textbf{99.17} & 99.03 \\ \hline
    Soil & 99.81 & 98.20 & 90.09 & 96.64 & \textbf{100.0} & 98.61 & 97.66 & 99.99 & 99.78 & 98.58 & 98.29 & 99.83 \\ \hline
    Water & \textbf{100.0} & 99.30 & 93.94 & 84.66 & 87.53 & 95.08 & 97.37 & 98.42 &  97.47 & 96.42 & 96.39 & 95.67 \\ \hline
    Residential & 87.59 & 98.97 & 87.53 & 78.78 & 77.00 & 91.46 & 91.93 & 84.32 & 86.43 & 89.18 & 91.60 & \textbf{93.53} \\ \hline
    Commercial & 73.22 & 98.01 & 93.02 & 45.49 & 49.77 & 78.02 & 94.88 & 80.80 & 83.25 & 83.90 & 83.33 & \textbf{96.94} \\ \hline
    Road & 81.21 & \textbf{95.94} & 86.12 & 76.05 & 72.19 & 92.26 & 88.57 & 78.82 & 82.61 & 86.58 & 92.05 & 92.17 \\ \hline
    Highway & 68.15 & 98.65 & 78.68 & 36.93 & 94.51 & 97.61 & 89.76 & 95.10 & 96.50 & 99.09 & 98.05 & \textbf{99.83} \\ \hline
    Railway & 89.85 & 87.38 & 81.65 & 64.05 & 85.59 & 93.82 & 95.44 & 95.82 & 93.46 & 92.34 & 92.48 & \textbf{95.95} \\ \hline
    Parking-lot-1 & 89.15 & 94.33 & 89.71 & 70.01 & 59.63 & 92.45 & 93.15 & 89.65 & 89.55 & 91.27 & 91.24 & \textbf{94.97} \\ \hline
    Parking-lot-2 & 92.28 & 98.95 & 85.49 & 98.81 & 80.06 & 95.86 & 82.75 & \textbf{97.75} & 92.98 & 91.22 & 92.87 & 89.74 \\ \hline
    Tennis-Court & \textbf{100.0} & 98.79 & 93.02 & 99.81 & \textbf{100.0} & 99.95 & 98.13 & 99.85 & 99.88 & \textbf{100.0} & \textbf{100.0} & 97.66 \\ \hline
    Running-Track & \textbf{100.0} & 62.16 & 88.33 & 96.41 & 99.57 & 99.45 & 99.05 & \textbf{100.0} & 99.97 & 98.63 & \textbf{100.0} & 99.09 \\ \hline \hline 
    \textbf{OA} & 86.30 & 93.21 & 86.50 & 75.65 & 82.14 & 93.92 & 93.67 & 92.88 & 92.85 & 92.73 & 94.30 & \textbf{96.92} \\ \hline
    \textbf{AA} & 88.68 & 93.22 & 85.25 & 79.34 & 84.19 & 94.65 & 94.03 & 94.10 & 93.79 & 93.51 & 94.96 & \textbf{96.73} \\ \hline
    \textbf{$\kappa$} & 85.14 & 92.63 & 85.85 & 73.73 & 80.70 & 93.43 & 93.16 & 92.30 & 92.27 & 92.14 & 93.84 & \textbf{96.67} \\ \hline
    \end{tabular}}
    \label{Tab5}
\end{table}

For the University of Houston dataset, CAT, MSSG, DBDA, SSFTT, LSFAT, and CT-Mixer demonstrated superior performance compared to SPRLT, CMT, S3L, and RIAN, with improvements of approximately 6-10\% in OA, AA, and $\kappa$ accuracy metrics. SS-Mamba surpassed these models by an additional 2\% across OA, AA, and $\kappa$ metrics. Our proposed model, MHSSMamba, further enhanced performance, exceeding SS-Mamba by 2.62\% in OA, 1.77\% in AA, and 2.83\% in $\kappa$ accuracy. These results underscore MHSSMamba's effectiveness across various learning frameworks. Transformer-based models, requiring higher learning rates and more epochs, showed lower performance on Houston data, likely due to dataset characteristics.

\begin{table}[!hbt]
\centering
\caption{Pavia University Classification Results over several SOTA methods.}
\resizebox{\columnwidth}{!}{\begin{tabular}{c|cccccccccccc} \hline 
    \textbf{Class} & RIAN & CAT & S3L & SPRLT & CMT & MSSG & DBDA & SSFTT & LSFAT & CT-Mixer & SS-Mamba & \textbf{Proposed} \\ \hline 
    Asphalt & 90.44 & 94.75 & 98.76 & 78.09 & 76.94 & 97.06 & 98.74 & 86.16 & 87.16 & 90.59 & 95.70 & \textbf{98.46} \\ \hline
    Meadows & 96.63 & 95.54 & 95.46 & 84.67 & 93.70 & 92.29 & 99.51 &9 4.50 & 95.05 & 94.62 & 94.05 & \textbf{99.73} \\ \hline
    Gravel & 93.50 & 99.83 & 87.75 & 53.47 & 81.10 & \textbf{99.97} & 90.81 & 93.94 & 93.95 & 94.39 & 99.61 & 91.89 \\ \hline
    Trees & 90.94 & 93.99 & 91.37 & 75.09 & 78.42& 97.12 & 92.74 & 88.97 & 92.76 & 84.61 & \textbf{98.92} & 96.73 \\ \hline
    Mental Sheets & 98.71 & 95.82 & 92.13 & 99.83 & \textbf{100.0} & \textbf{100.0} & 99.53 & 98.95 & 98.85 & 99.18 & \textbf{100.0} & \textbf{100.0} \\ \hline
    Bare Soil & 98.82 & 97.32 & 97.10 & 52.40 & 64.24 & 99.40 & 90.96 & 96.07 & 99.19 & \textbf{99.53} & 99.19 & 98.05 \\ \hline
    Bitumen & 98.13 & 96.58 & 96.47 & 78.74 & 96.27 & \textbf{100.0} & 93.80 & 99.58 & 99.08 & 99.34 & 99.93 & 96.39 \\ \hline
    Bricks & 93.95 & 95.37 & 86.25 & 84.35 & 58.58 & \textbf{98.99} & 89.83 & 86.63 & 91.74 & 97.10 & 98.50 & 88.37 \\ \hline
    Shadow & 95.24 & 92.18 & 89.27 & 99.60 & 98.16 & 99.47 & 96.82 & 95.54 & 95.49 & 93.95 & \textbf{99.96} & 98.10 \\ \hline \hline 
    \textbf{OA} & 95.25 & 95.05 & 94.64 & 78.23 & 83.28 & 95.79 & 95.87 & 92.61 & 94.06 & 94.33 & 96.40 & \textbf{96.41} \\ \hline
    \textbf{AA} & 95.15 & 93.01 & 92.78 & 78.47 & 83.06 & 98.25 & 94.75 & 93.37 & 94.81 & 94.81 & \textbf{98.43} & 97.62 \\ \hline
    \textbf{$\kappa$} & 93.55 & 94.70 & 93.54 & 71.55 & 77.74 & 94.54 & 94.58 & 90.29 & 92.22 & 92.60 & 95.31 & \textbf{96.85} \\ \hline
    \end{tabular}}
    \label{Tab4}
\end{table}

For the Pavia University dataset, traditional Spatial-Spectral models like SPRLT and CMT were less effective compared to RIAN, CAT, and S3L, which showed significant improvements of 12\% in OA, AA, and $\kappa$ metrics. RIAN, CAT, and S3L also outperformed DBDA, SSFTT, LSFAT, and CT-Mixer by 1-3\% in OA, AA, and $\kappa$ accuracy. MSSG surpassed RIAN, CAT, and S3L with gains of 0.50\% in OA, 3.10\% in AA, and 1.05\% in $\kappa$ accuracy. The SS-Mamba model outperformed MSSG by 1.0\% in OA, AA, and $\kappa$ accuracy. Lastly, MHSSMamba showed slight improvements over SS-Mamba in OA and $\kappa$ accuracy, while SS-Mamba achieved slightly better results in AA.

\section{Computational Complexity}

The computational complexity of the proposed model can be analyzed by evaluating each of its primary components, for instance, the token generation layer, which generates the spatial-spectral tokens using the dense layer, has a complexity of $O(B \times H \times W \times C \times out\_Channels)$. The Multi-head self-attention layer involves matrix multiplications and attention score calculations, resulting in a complexity of $O(B \times L \times embed\_Dim^2 + B \times num\_heads \times L^2 \times head\_dim)$. The feature enhancement layer responsible for enhancing features using dense layers and element-wise operations, contributed $O(B \times L \times out\_Channels^2)$. Finally, the SSM layer which updates the state through dense layers over $T$ timesteps has a complexity of $O(T \times B \times State\_dim^2)$. Combining all layers, the overall complexity of the MHSSMamba is $O(B \times H \times W \times C \times out\_Channels^2 + B \times L \times embed_dim^2 + B \times num\_heads \times L^2 \times head\_dim + B \times L \times out\_Channels^2 + T \times B \times state\_dim^2)$, reflecting the cumulative computational demands of token generation, attention mechanisms, feature enhancement, and state updates. Here we also discuss the best, average, and worst-case scenarios of the proposed model. 

\textbf{Best Case:} When the model operates with minimal features and token sizes, and there is no need for extensive processing, the complexity of the token generation is $O(B \times H \times W \times C)$, where $B$ is the batch size, $H, ~W$ are the spatial dimensions, and $C$ is the number of bands. The multi-head self-attention under the minimal head and embedding sizes has a complexity of $O(B \times L \times embed\_dim)$, and the SSm assuming minimal state dimension and timesteps length operates with $(T \times B \times state\_dim)$. \textbf{Average Case:} With the typical sizes for token generation, the complexity will be $O(B \times H  \times W  \times C  \times out\_Channels)$. the multi-head self-attention considering the standard attention heads and embeddings sizes, results in $O(B  \times L  \times embed\_dim^2 + B  \times num\_heads  \times L^2  \times head\_dim)$. For the feature enhancement module, the complexity is $O(B  \times L  \times out\_channels^2)$, while the SSm operates at $O(T  \times B  \times state\_dim^2)$. \textbf{Worst Case:} In scenarios with maximal feature and token sizes, the complexity of the token generation can reach $O(B \times H \times W \times C \times out\_Channels)$. The multi-head self-attention complexity in the worst case is $O(B \times L^2 \times emded\_dim^2)$ due to the large attention scores and weights. The feature enhancement layers scale to $O(B \times L \times  out\_Channels^2)$, and the SSM could require $O(T \times B \times state\_dim^2)$ in the most extensive cases.

\section{Conclusions}

This paper introduces the Multihead Attention-Based Mamba (MHSSMamba) architecture for HSIC. Tested on benchmark datasets, MHSSMamba was evaluated using Overall Accuracy (OA), Average Accuracy (AA), and the kappa coefficient ($\kappa$). It outperformed state-of-the-art methods, achieving 97.82\% OA, 96.41\% AA, and 96.85\% $\kappa$ on the Pavia University dataset, and 96.92\% OA, 96.41\% AA, and 97.62\% $\kappa$ on the University of Houston dataset. These results highlight the model's robustness and effectiveness in capturing spectral-spatial features. The superior performance is due to its advanced multi-head attention mechanisms, which enhance spectral-spatial information extraction. 

\section*{Disclosure statement}

The authors declare no conflict of interest exists.

\section*{Data Availability}

The data used in this study is publicly available and can be accessed from the corresponding data repository \href{https://www.ehu.eus/ccwintco/index.php/Hyperspectral_Remote_Sensing_Scenes}{Hyperspectral Datasets}.





\footnotesize
\bibliographystyle{IEEEtran}
\bibliography{IEEEabrv,Sam}
\end{document}